\documentclass{article}
\usepackage{spconf,amsmath,epsfig}

\usepackage[]{graphicx}
\usepackage{caption}
\usepackage{subfig} 
\usepackage{amsmath}
\usepackage{amssymb}
\usepackage{epsfig}
\usepackage{cite}
\usepackage{color}
\usepackage{balance}
\usepackage{pgfplots}
\usepackage{amsmath}
\usepackage{amsfonts} 
\usepackage{graphicx}
\usepackage{pdfpages}
\usepackage[T1]{fontenc} 
\usepackage{array}
\usepackage{url}

\usepackage{color}
\usepackage{wrapfig}
\usepackage{lipsum}
\usepackage[hidelinks]{hyperref}
\usepackage{multirow}
\usepackage{tabularx}
\usepackage{gensymb} 
\usepackage{tikz}
\usepackage{color}

\newcolumntype{L}{>{\arraybackslash}m{3cm}} 
\newcolumntype{M}{>{\centering\arraybackslash}m{1.2cm}} 

\title{DESIGN OF LOW-COST, COMPACT AND WEATHER-PROOF\\
WHOLE SKY IMAGERS for High-Dynamic-Range captures}

\name{Soumyabrata Dev,$^{1}$ Florian M. Savoy,$^{2}$ Yee Hui Lee,$^{1}$ Stefan Winkler$^{2}$\thanks{This research is funded by the Defence Science and Technology Agency (DSTA), Singapore.}\thanks{Send correspondence to Y.H.\ Lee, E-mail: eyhlee@ntu.edu.sg.}}
\address{$^{1}$~School of Electrical and Electronic Engineering, Nanyang Technological University (NTU), Singapore \\ 
$^{2}$~Advanced Digital Sciences Center (ADSC), University of Illinois at Urbana-Champaign, Singapore \\
}

\begin{document}

\maketitle

\begin{abstract}
Ground-based whole sky imagers are popular for monitoring cloud formations, which is necessary for various applications. We present two new Wide Angle High-Resolution Sky Imaging System (WAHRSIS) models, which were designed especially to withstand the hot and humid climate of Singapore. The first uses a fully sealed casing, whose interior temperature is regulated using a Peltier cooler. The second features a double roof design with ventilation grids on the sides, allowing the outside air to flow through the device. Measurements of temperature inside these two devices show their ability to operate in Singapore weather conditions. Unlike our original WAHRSIS model, neither uses a mechanical sun blocker to prevent the direct sunlight from reaching the camera;  instead they rely on high-dynamic-range imaging (HDRI) techniques to reduce the glare from the sun.
\end{abstract}


\section{Introduction}
\label{sec:intro}
There is a growing interest in using whole sky imagers for monitoring clouds. These imagers are used in geoscience and remote sensing for applications such as solar energy prediction \cite{solar_irr_pred}, attenuation analysis in satellite-to-ground communication links \cite{cloud_models}, estimation of cloud fraction and other microphysical properties \cite{cloud_fraction}. Such imagers complement the information obtained from satellite images \cite{optical_sensing}. They provide a higher resolution of a specific area and an upwards pointing view, which facilitates sky/cloud segmentation \cite{ICIP1_2014,ICIP2015a}.  

Commercially available sky imagers (TSI-440, TSI-880) manufactured by Yankee Environmental Systems\footnote{~\url{http://www.yesinc.com/}} are used by a number of research groups \cite{Long}. However, these devices are expensive, provide poor image resolution, and offer only limited control flexibility. This motivated us to build our own imagers using off-the-shelf components.  Other researchers have done likewise \cite{FirstWSI,WSI-history}.

In an earlier publication\cite{WAHRSIS}, we proposed the initial Wide Angle High-Resolution Whole Sky Imaging System (WAHRSIS) model. It uses a mechanical sun blocker that follows the sun to reduce glare in the images. Its mounting structure and motors reduce the visible sky region quite significantly (cf.\ Figure \ref{fig:sun-blocking}a).  After extensive usage of this model, two additional operational problems became apparent: (i) Since the motors moving the sun blocker are exposed to rain, they have started to rust and do not work anymore; (ii) The enclosure heats up when placed in the sun, and the temperature inside often exceeds the recommended operating range of the electronic devices.

In this paper, we introduce two improved WAHRSIS models that were designed to solve all of the above issues. We achieve a considerable reduction in size without compromising on performance. Instead of using a conventional sun blocker to reduce sun glare in the image, we employ High-Dynamic-Range Imaging (HDRI) techniques~\footnote{The source code for the several simulations in this paper is available online at \url{https://github.com/Soumyabrata/HDRCaptures}.}.

The rest of the paper is organized as follows: Section 2 discusses the design of both WAHRSIS models. Temperature measurements inside the devices are carried out in Section 3. Section 4 presents the high-dynamic-range imaging procedure. Section 5 concludes the paper.

\section{Design}
A sky imager essentially comprises a DSLR camera with a fish-eye lens, which are placed inside a weather-proof casing with a transparent dome. A single board computer instructs the camera to take pictures at user-defined intervals, stores the captured images in an on-board hard drive, and may even send them to a server. 

In this section, we detail the mechanical design of the new WAHRSIS models. They were designed to withstand the hot and humid climate of Singapore and run for extended periods of time without interruption or human intervention. 

Two strategies were implemented. The first model is a fully sealed imager with a Peltier cooler to control the internal temperature. The second model has a double roof, which isolates the casing from direct sunlight. It also features ventilating grids, allowing the outside air to flow through the device. The compact size of both models minimizes the volume of air to control and also enhances their portability. The transformers and power strip are placed outside the main device in another polystyrene box as they are less sensitive to high temperatures and also generate heat themselves.

The total component cost of our new WAHRSIS models is less than US\$~2000 each, as specified in Table \ref{tab:WSI-cost}. 

\begin{table}[htb]
\small 
\centering
\begin{tabular}{ lcc }
  \hline 
  \textbf{Items} & \textbf{Sealed model} & \textbf{Ventilated model}  \\
  \hline 
  DSLR camera & 677 & 677 \\ 
  Fish-eye lens & 733  & 733  \\ 
  Single-board PC & 68 & 68  \\ 
  Mechanical items & 212  & 221 \\ 
  Peltier cooler &  65 & - \\ 
  Miscellaneous & 140  & 70  \\ \hline 
  \textbf{Total cost} & \textbf{\$1895} & \textbf{\$1769} \\ \hline 
\end{tabular}
\caption{Breakdown of the  costs of the sealed and ventilated WAHRSIS models (in US\$). The mechanical items consist of the frame, box, screws etc., whereas miscellaneous items include the extension power cord, the power adapter etc.}
\label{tab:WSI-cost}
\end{table}

\subsection{Sealed WAHRSIS Model}
\begin{figure}[htb]
\centering
\begin{tikzpicture}
\begin{scope}[xshift=1.5cm]
    \node[anchor=south west,inner sep=0] (image) at (0,0) {\includegraphics[width=0.35\textwidth]{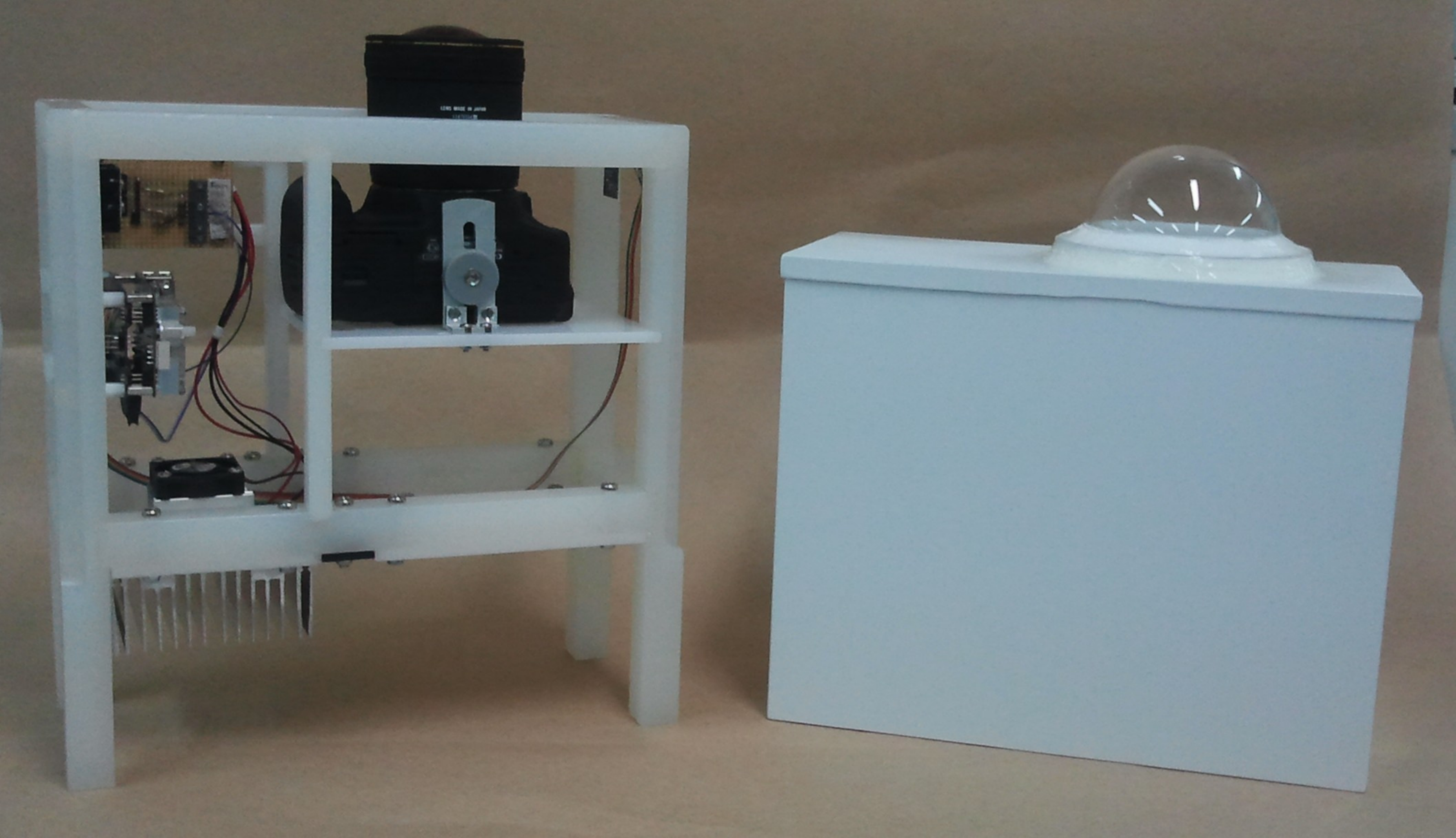}};
    \begin{scope}[x={(image.south east)},y={(image.north west)}]
        \draw[->,black,dashed](0.3,0.9) -- (1.1,0.9) node[anchor=west,black] {(a)};
        \draw[->,black,dashed](0.8,0.77) -- (1.1,0.77) node[anchor=west,black] {(b)};
        \draw[->,black,dashed](0.1,0.6) -- (1.1,0.6) node[anchor=west,black] {(c)};
        \draw[->,black,dashed](0.2,0.3) -- (1.1,0.3) node[anchor=west,black] {(e)};
        \draw[->,black,dashed](0.8,0.45) -- (1.1,0.45) node[anchor=west,black] {(d)};
    \end{scope}
\end{scope}
\end{tikzpicture}
\caption{Sealed WAHRSIS design. (a) DSLR camera, (b) dome, (c) ODROID board, (d) casing, (e) thermoelectric cooler with fans.}
\label{fig:W3-design}
\end{figure}
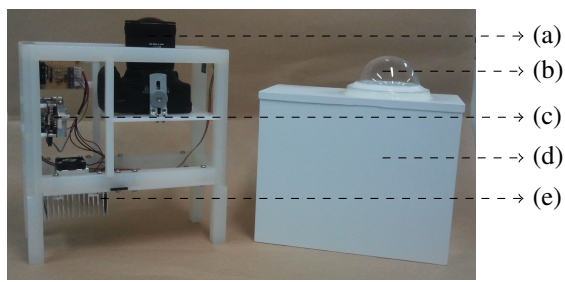

The sealed WAHRSIS model consists of a hermetically-sealed plastic box with sunlight-reflecting paint on it, as shown in Fig. \ref{fig:W3-design}. A thermoelectric Peltier cooler is placed at the bottom of the casing. When electricity is provided, the side facing inside the box gets cold while the outside face gets hot. Fans are placed at both sides to dissipate the cold or heat produced. The on-board computer controls the cooler operation using temperature and humidity sensors. 

\subsection{Ventilated WAHRSIS Model}
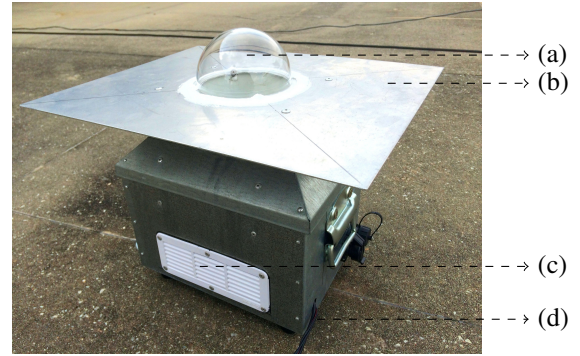
\begin{figure}[htb]
\centering
\begin{tikzpicture}
\hspace{0.8cm}
\begin{scope}[xshift=1cm]
    \node[anchor=south west,inner sep=0] (image) at (0,0) {\includegraphics[width=0.35\textwidth]{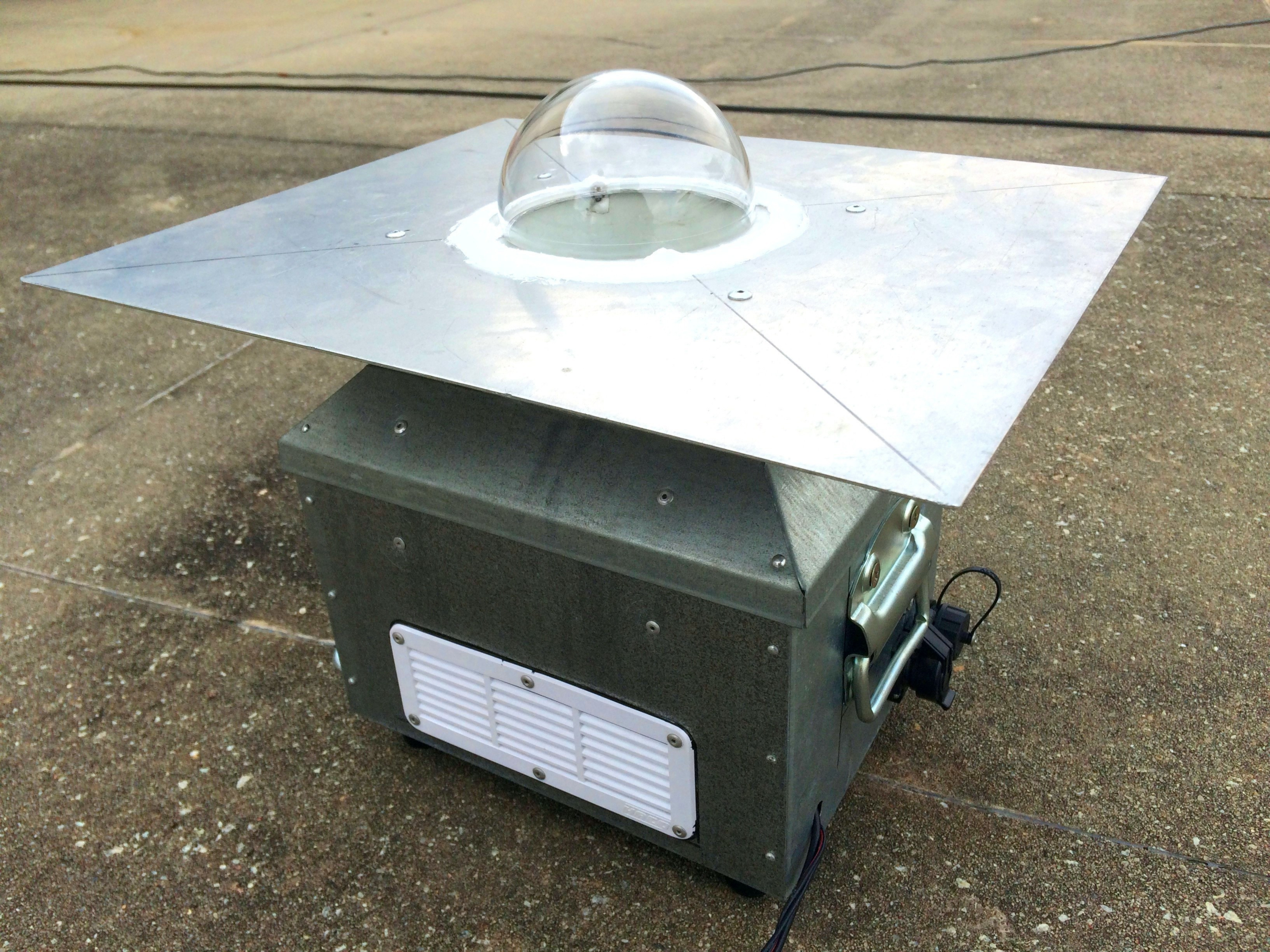}};
    \begin{scope}[x={(image.south east)},y={(image.north west)}]
        \draw[->,black,dashed](0.5,0.85) -- (1.1,0.85) node[anchor=west,black] {(a)};
        \draw[->,black,dashed](0.8,0.77) -- (1.1,0.77) node[anchor=west,black] {(b)};
        \draw[->,black,dashed](0.4,0.25) -- (1.1,0.25) node[anchor=west,black] {(c)};
        \draw[->,black,dashed](0.65,0.1) -- (1.1,0.1) node[anchor=west,black] {(d)};
    \end{scope}
\end{scope}
\end{tikzpicture}
\caption{Ventilated WAHRSIS design. (a) Dome, (b) shelter, (c) air vent cover, (d) casing. }
\label{fig:version2}
\end{figure}

The ventilated WAHRSIS model consists of a casing made of galvanized steel, as shown in Fig.~\ref{fig:version2}. This material reflects the sun light well and prevents rusting and corrosion. Ventilation grids are placed on each side, allowing ambient air to flow inside while preventing rainwater from entering. A metallic sheet is placed a few centimeters on top of the box, along with the transparent dome for the camera lens. The air in between acts as insulator, which protects the case from being heated by direct sunlight. 

\section{Temperature Control}
Singapore has a typical tropical climate, i.e.\ hot and humid conditions throughout the year, with average low/high temperatures of 25$^{\circ}$C/33$^{\circ}$C, respectively. A box placed outside in the sun tends to heat up quickly, so we need to make sure the maximum operating temperature of the camera, lens, and the other electronic components of 40$^{\circ}$C is not exceeded. 

For the sealed box, we set two distinct thresholds for the operation of the Peltier cooler: we switch it on when the inside temperature reaches 37$^{\circ}$C, and off again when it falls below 32$^{\circ}$C. 

\begin{figure}[htb]
\begin{center}
\includegraphics[width=0.45\textwidth]{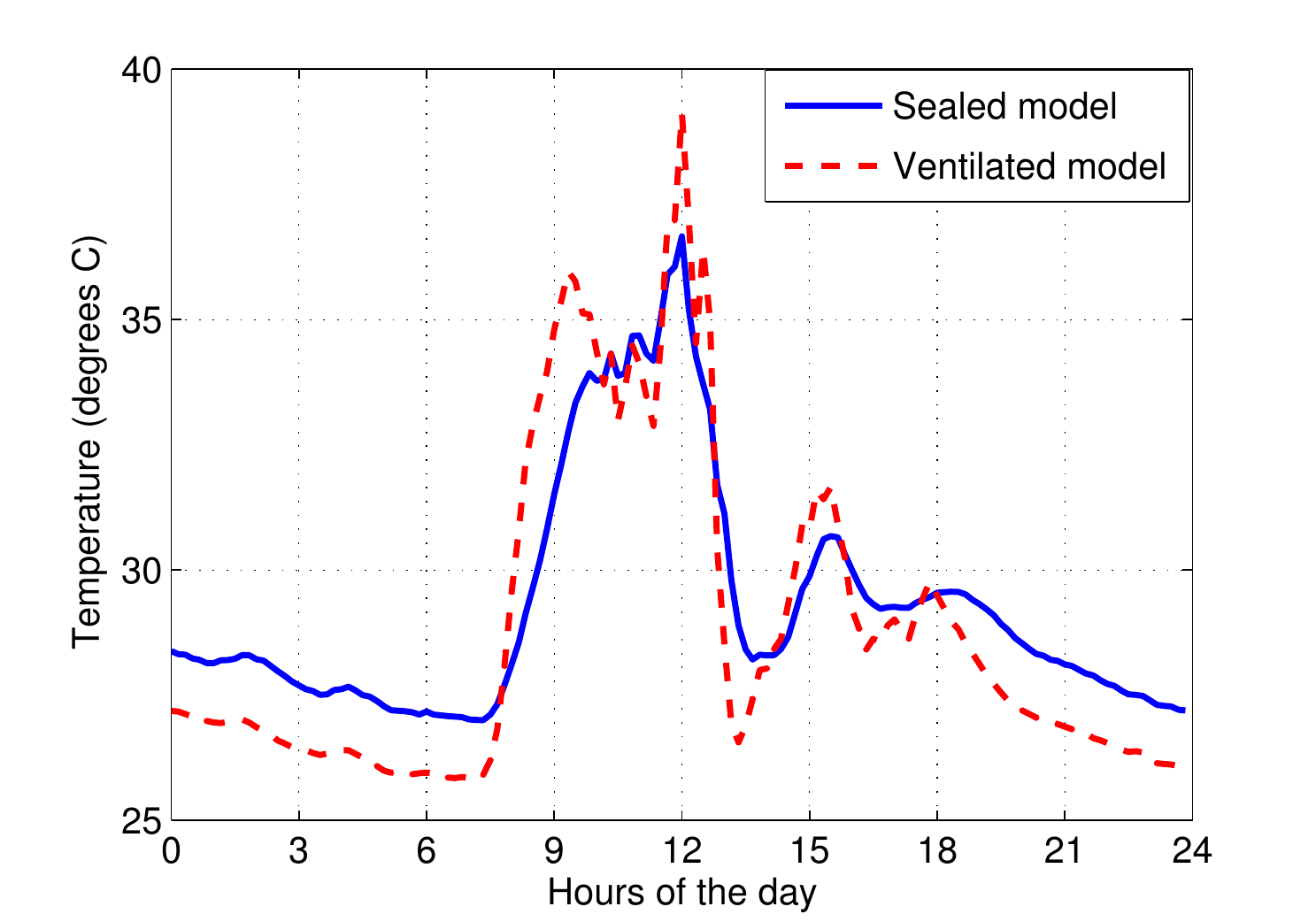}
\caption{Temperature recorded over the course of a day inside both models.
\label{fig:meas-rec}}
\end{center}
\end{figure}

In order to investigate the efficacy of the two designs, both WAHRSIS models were placed on a rooftop at the Nanyang Technological University of Singapore. We measured the temperature inside the devices on a typical day in March 2015 with several hours of sunshine. Fig.~\ref{fig:meas-rec} shows the results.

In both models, the temperature never exceeds 40$^{\circ}$C. During daytime, the interior temperature remains lower in the sealed model than in the ventilated model because of its active cooling. Temperature variations are also smoother in the sealed box as a result of the controlled environment. Yet the grids of the ventilated model are also surprisingly effective at passive cooling. An important point is that we never noticed condensation inside the boxes. We thus conclude that both models can work well in tropical regions.

\section{High Dynamic Range Imaging}
The new WAHRSIS models presented here do not have a physical sun blocking mechanism. Instead, we use HDRI techniques to reduce the over-exposure resulting from sun glare. This avoids occlusions from the sun-blocker and its support structure in the captured images (which significantly complicate subsequent computer vision tasks), while at the same time keeping the number of saturated pixels in the circumsolar region to a minimum. 

The imaging system of our WAHRSIS models consists of a Canon EOS 600D digital SLR camera with a sensor resolution of $5184\!\times\! 3456$ pixels and a Sigma 4.5mm F2.8 EX DC circular fisheye lens that captures nearly $180$ degrees of the sky hemisphere. Both imagers use an ODROID-C1 single board computer; with its 1.5 GHz processor and 1 GB RAM, this board is powerful enough to perform on-board image processing tasks. 

\begin{figure}[htb]
\begin{center}
\subfloat[$-3$~EV]{\includegraphics[height=1in]{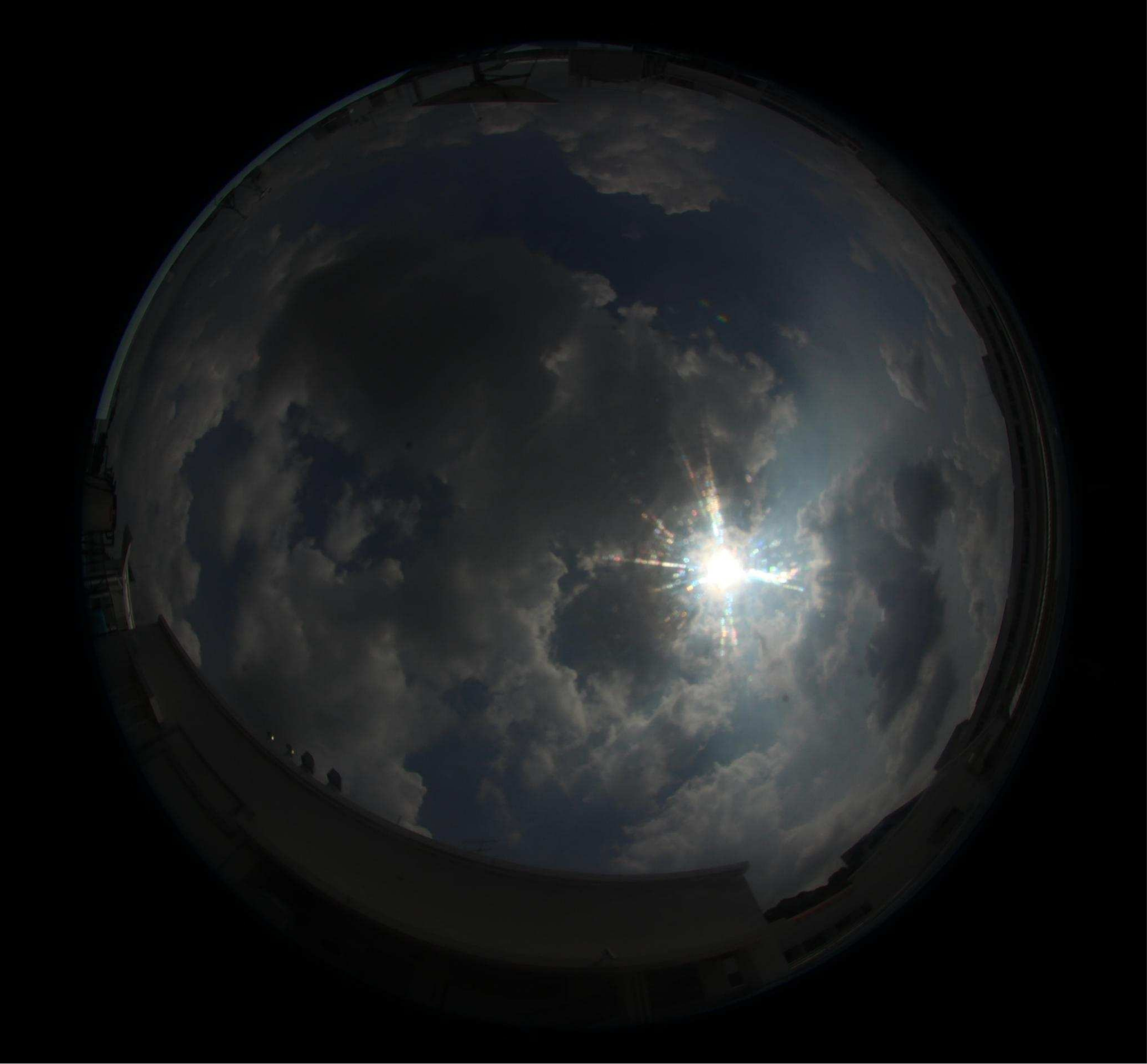}}
\subfloat[$-1$~EV]{\includegraphics[height=1in]{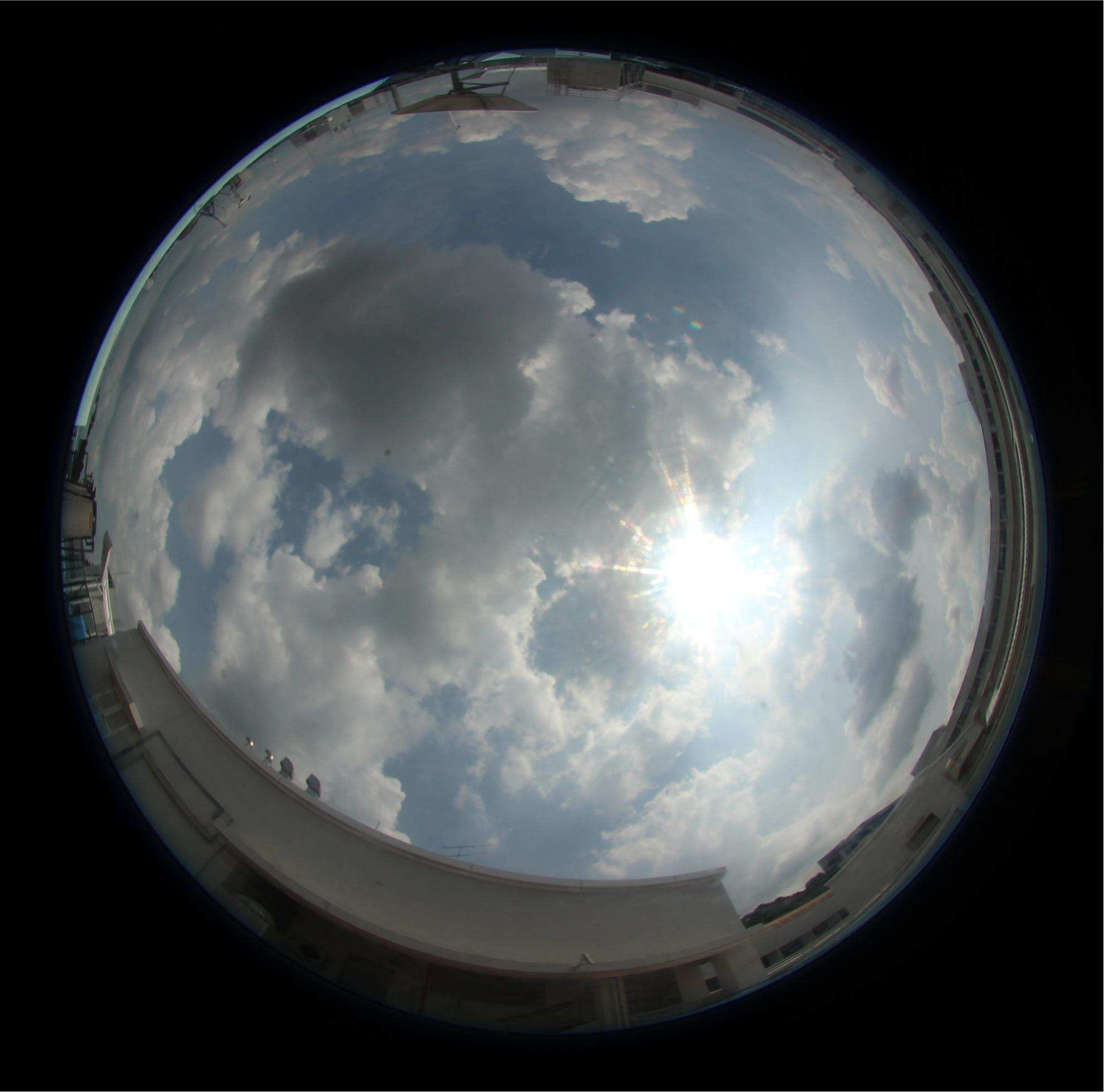}}
\subfloat[$+1$~EV]{\includegraphics[height=1in]{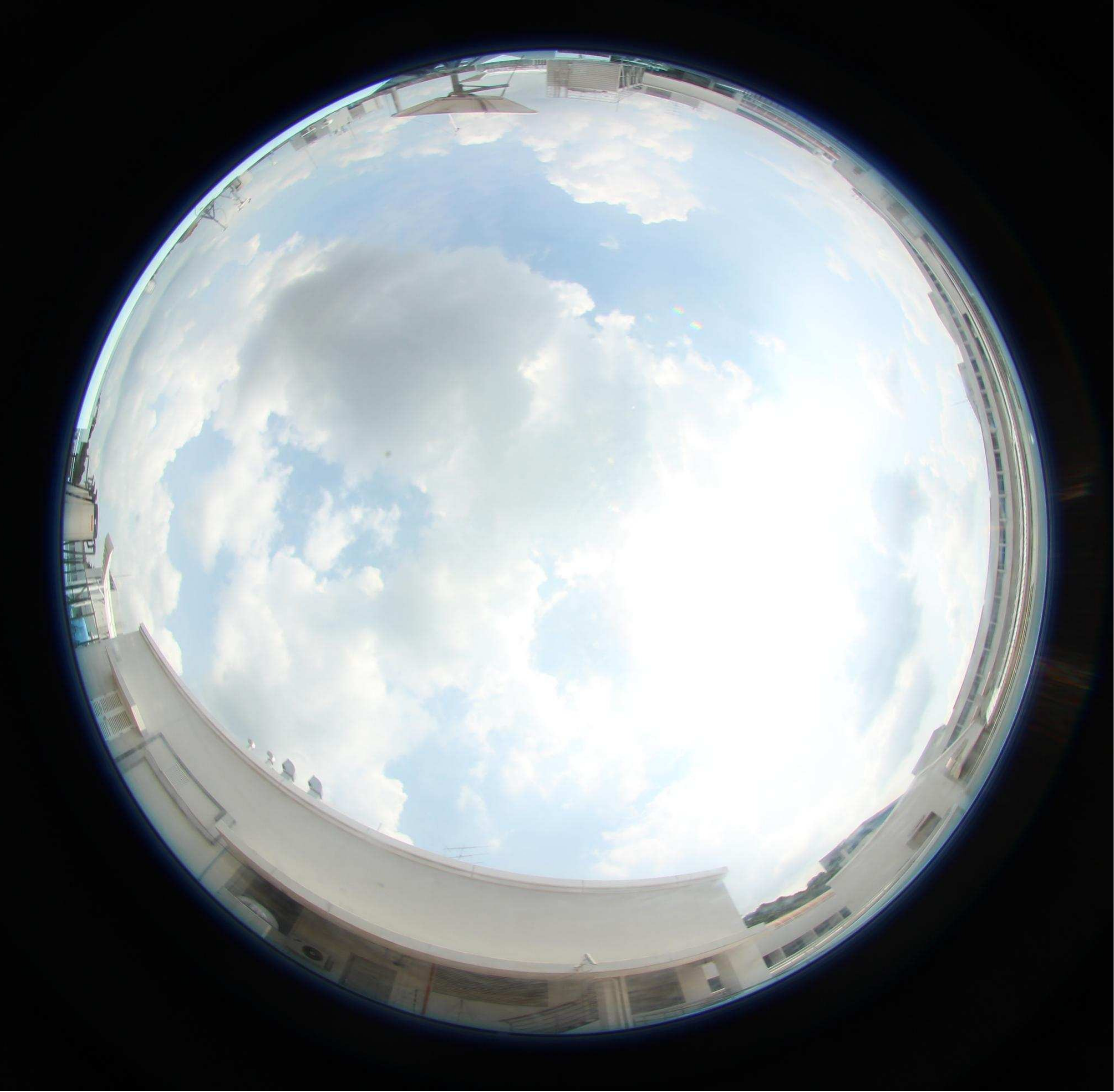}}
\caption{Low-dynamic-range (LDR) images captured at different exposure levels.
\label{fig:LDR-images}}
\end{center}
\end{figure}

The camera offers an automatic exposure bracketing (AEB) mode, which allows us to capture three (or more) images in burst mode with different exposure settings. We put the camera in aperture-priority mode to fix the aperture at the smallest available setting, and consequently the AEB function will vary the shutter speed only. For sky images we observed that the default exposure is usually somewhat too bright, so we choose $\{-3,-1,+1\}$ exposure value (EV) offsets for the bracketing.  Fig.\ \ref{fig:LDR-images} shows three Low-Dynamic-Range (LDR) images captured in this fashion.

\begin{figure}[htb]
\begin{center}
\includegraphics[width=0.45\textwidth]{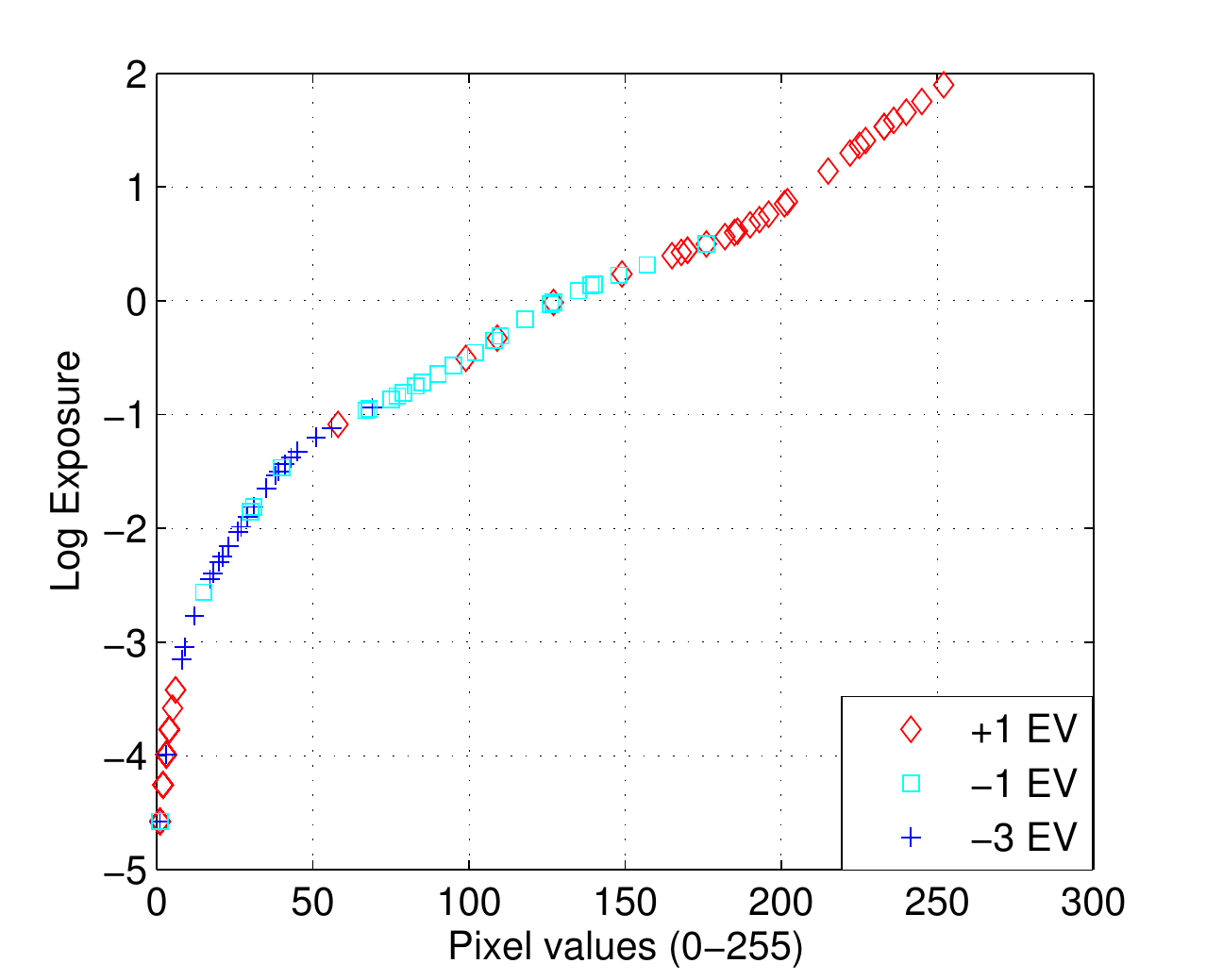}
\caption{Response curve of the HDRI system, computed from three LDR images with different exposure settings. 
}
\label{fig:response-RGB}
\end{center}
\end{figure}

Subsequently, we use the algorithm by Debevec and Malik \cite{DebevecHDR} to create a High-Dynamic-Range (HDR) radiance map from the three LDR images. As their exposure times are known, we can compute the camera response curve of the imaging system using samples obtained from the different LDR images. This is shown in Fig.~\ref{fig:response-RGB}. The dynamic range of the resulting  HDR radiance map is $15$ bits, which represents a significant improvement over the 8-bit  LDR images. 

\begin{figure}[htb]
\begin{center}
\subfloat[With sun blocker]{\includegraphics[height=0.32\columnwidth]{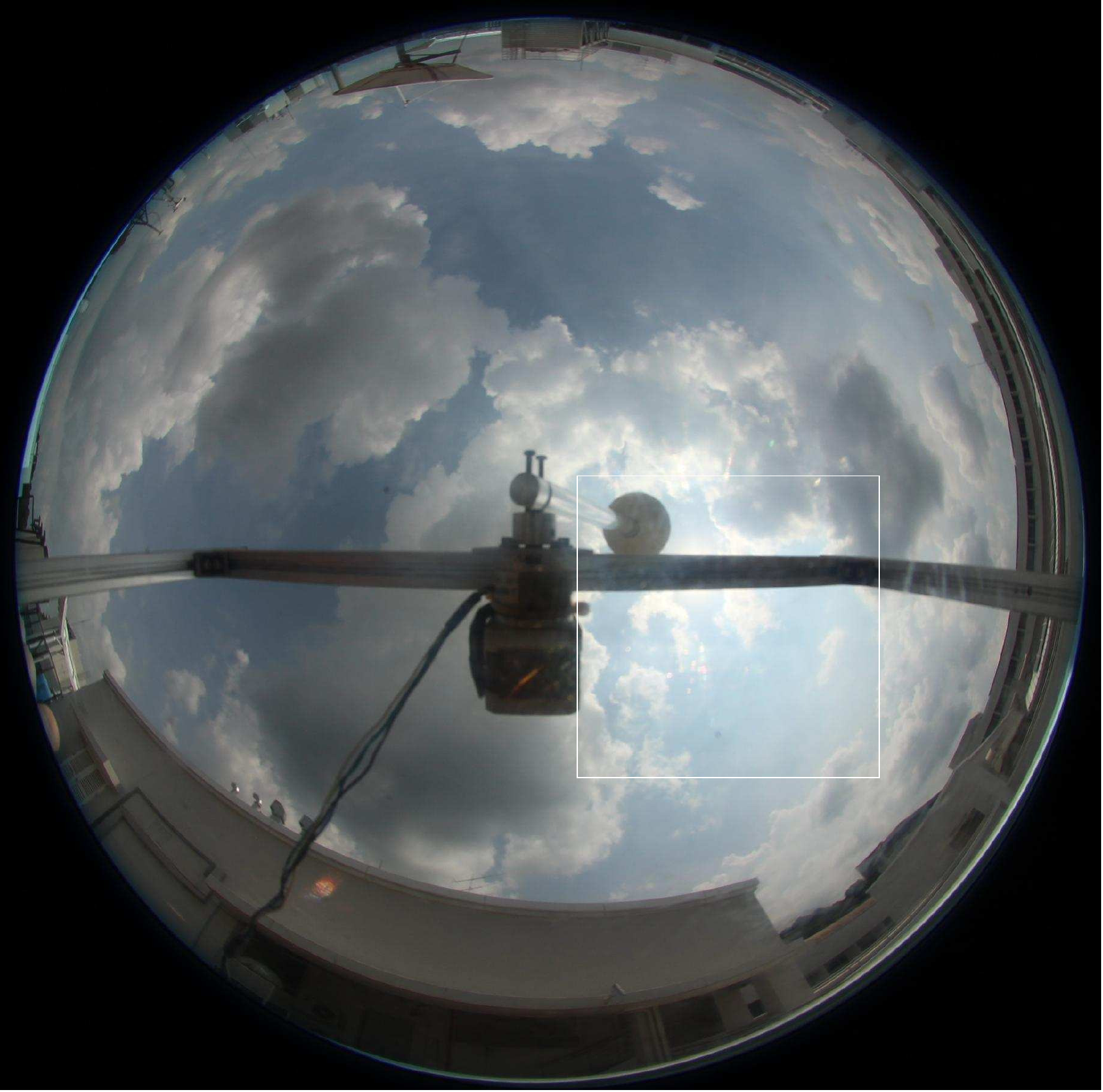}}
\subfloat[Auto exposure]{\includegraphics[height=0.32\columnwidth]{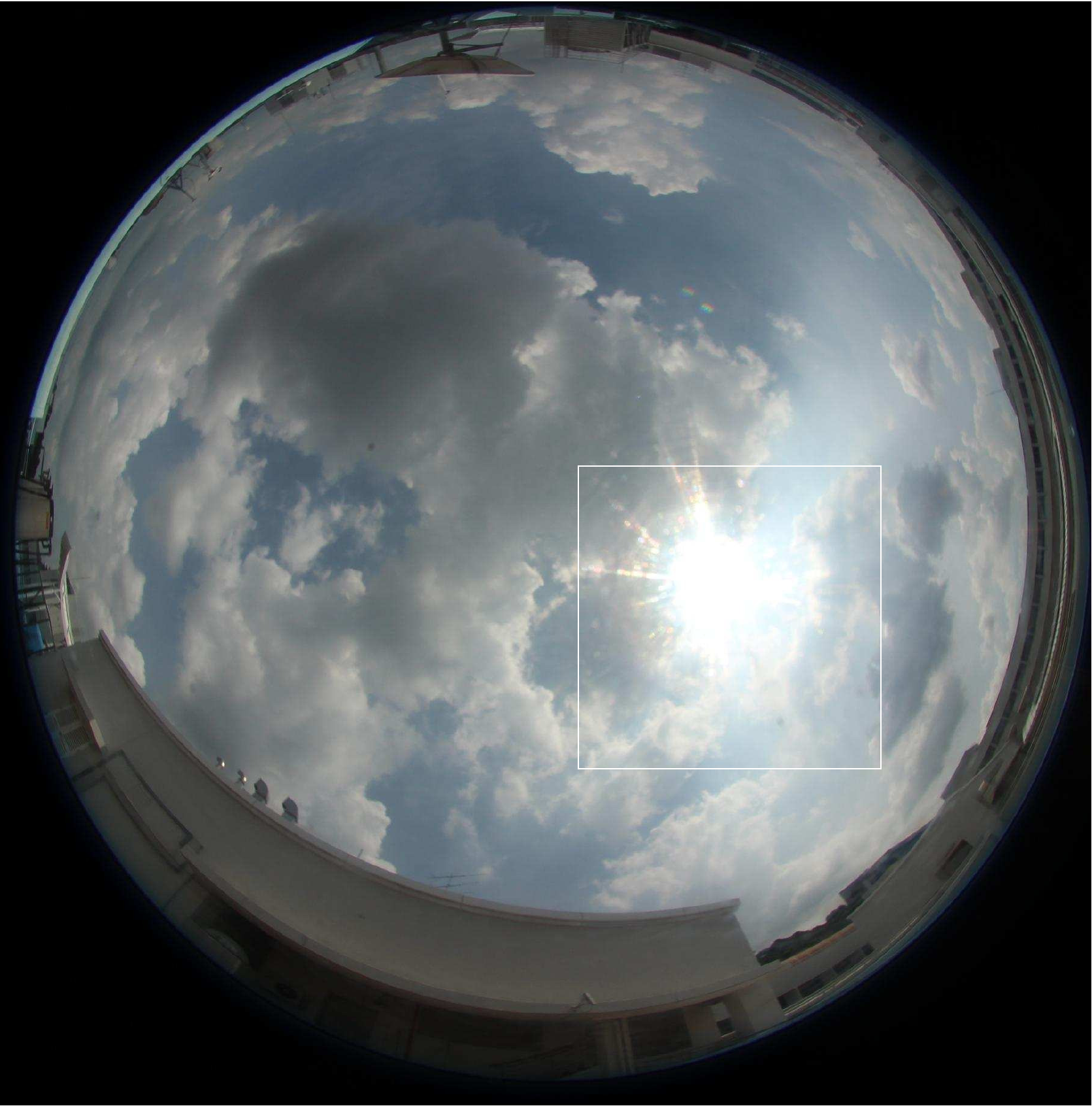}}
\subfloat[With HDR imaging]{\includegraphics[height=0.32\columnwidth]{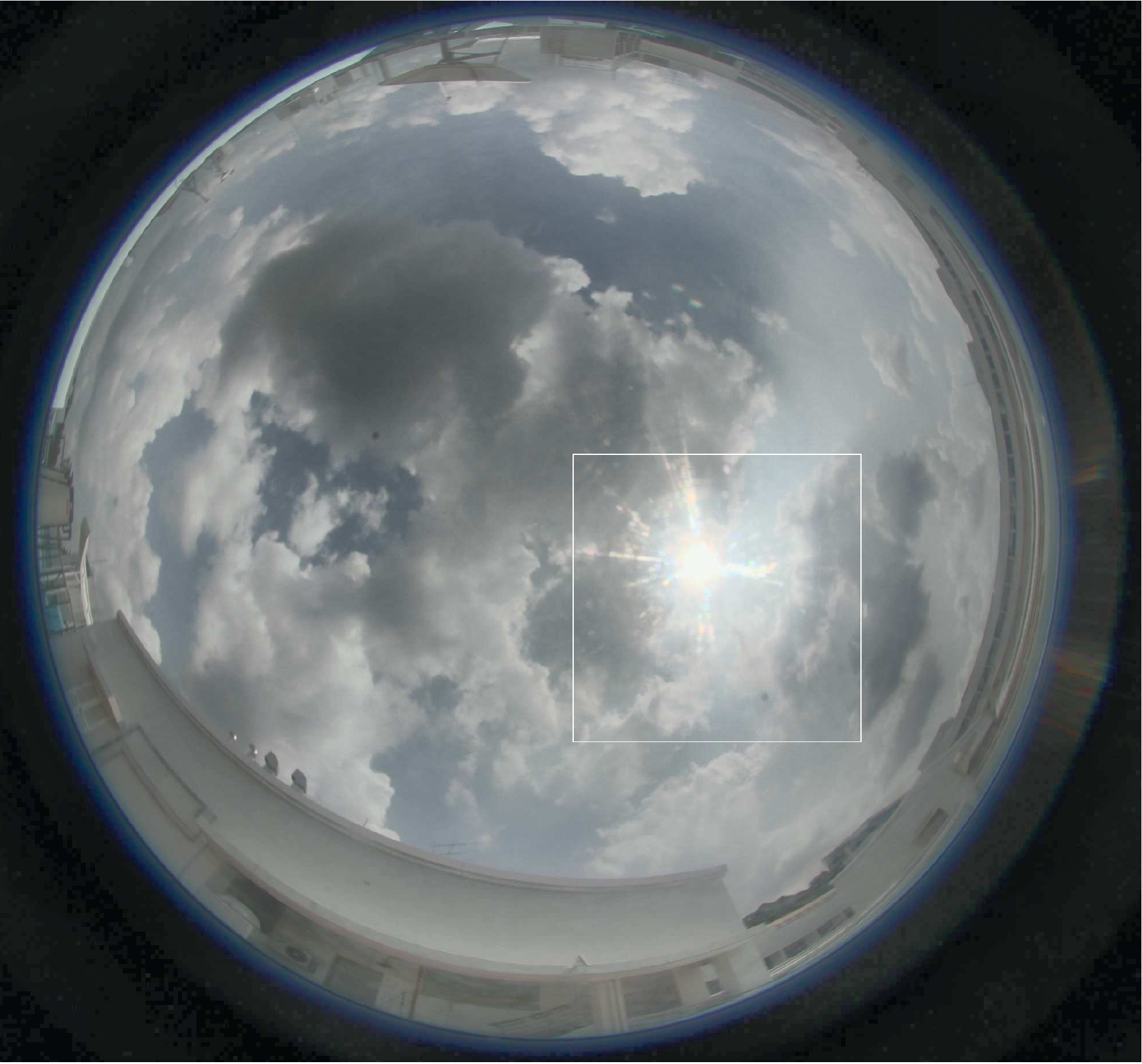}}\\
\vspace{-2.5mm}
\subfloat{\includegraphics[height=0.33\columnwidth]{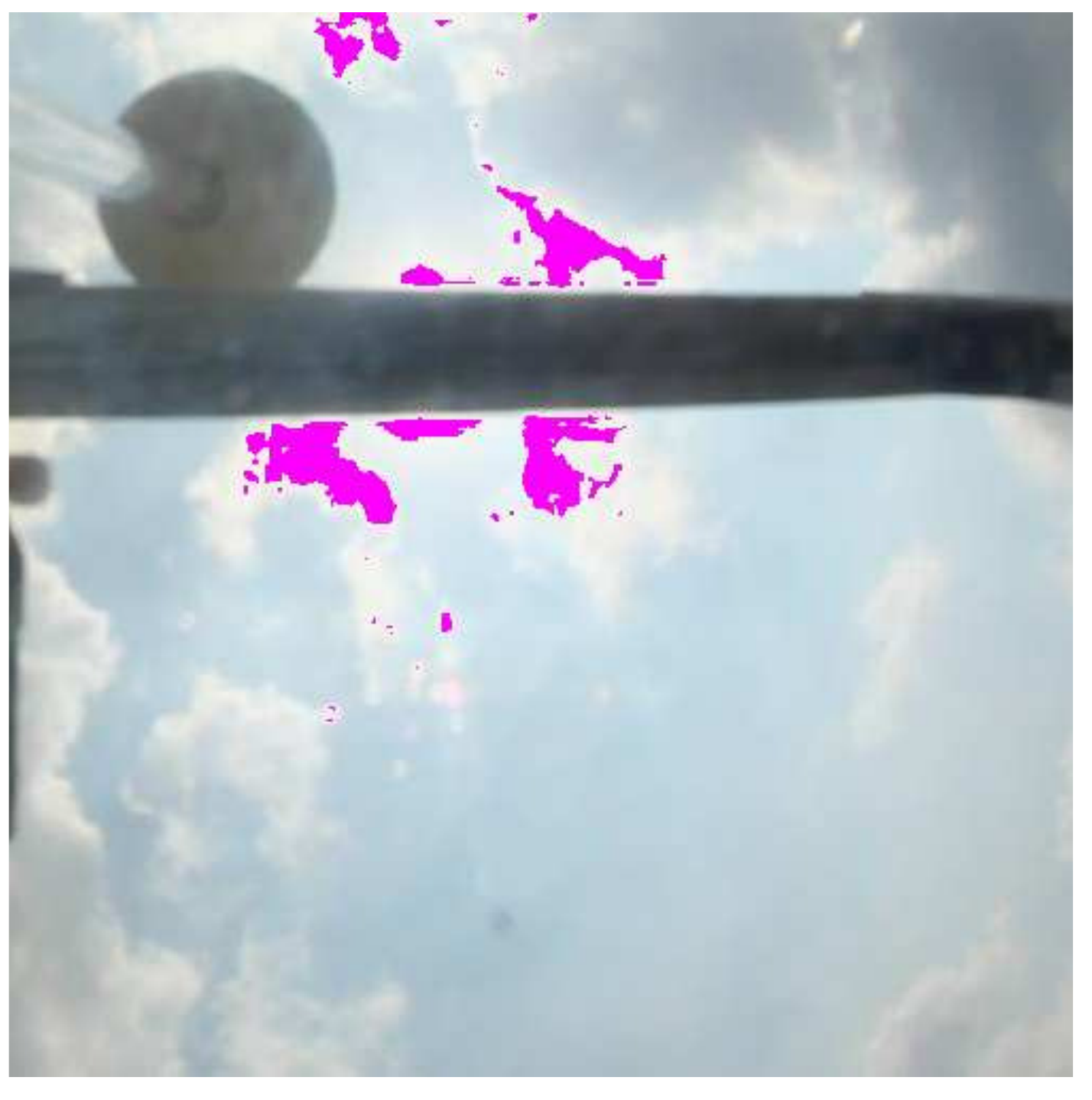}}
\subfloat{\includegraphics[height=0.33\columnwidth]{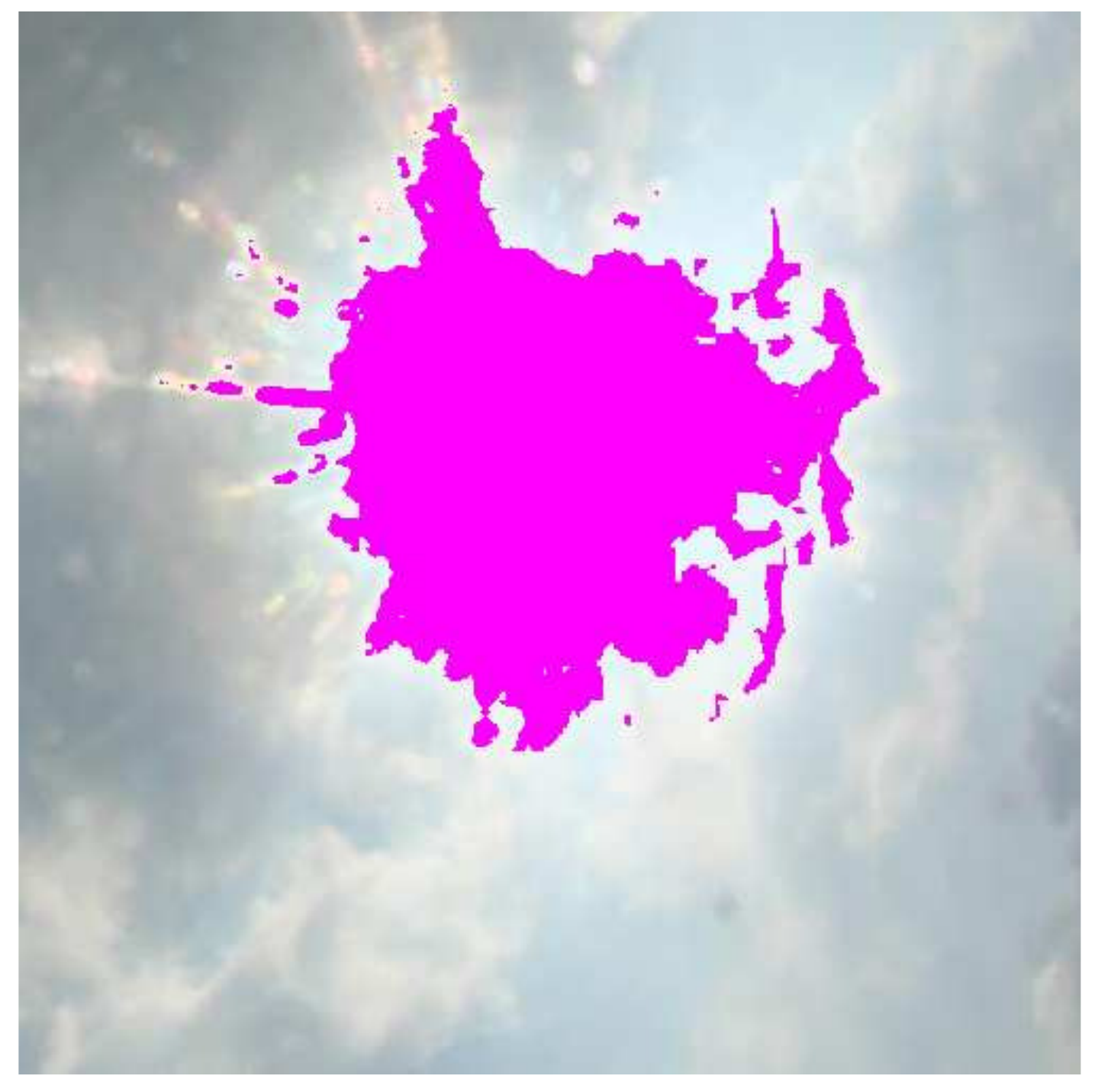}}
\subfloat{\includegraphics[height=0.33\columnwidth]{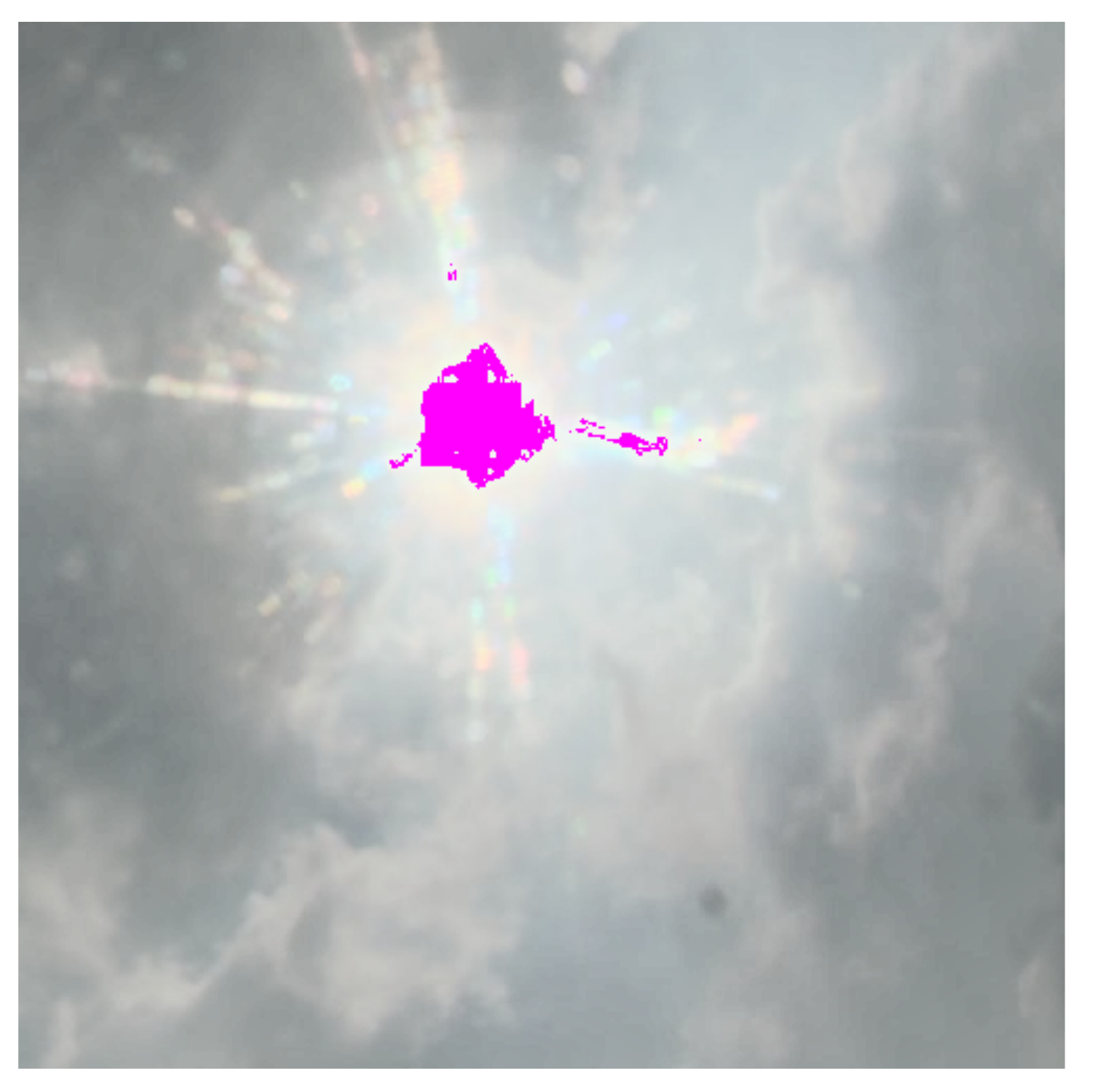}}
\caption{Effect of sun glare with different strategies.  The white rectangle outlines the circumsolar region, which is enlarged in the bottom row.  Saturated pixels are shown in pink.
\label{fig:sun-blocking}}
\end{center}
\end{figure}

In order to visualize the HDR image, we convert the high-dynamic-range radiance map to an LDR image via tone-mapping using an adaptive equalization process. Fig.~\ref{fig:sun-blocking}c shows the resulting image, along with those obtained from a sky imager with a mechanical sun blocker and a normal single-exposure image. In order to visualize the efficacy of our proposed approach, we zoom in on the circumsolar region in each of the three images to look at the amount of detail visible around the sun. Although the sun-blocker is effective at reducing sun glare, the occlusion is significant. Without a sun-blocker, the number of saturated pixels is very high in a normal single-exposure image. Fusing multiple exposures into an HDR image is indeed a good solution for minimizing the effects of over-exposure.

\section{Conclusions}
We introduced two new models of WAHRSIS, low-cost whole sky imagers built with off-the-shelf components. One features a sealed casing with thermoelectric cooling, while the other uses a ventilated casing. Measurements and extensive outdoor testing have proven their ability to maintain temperatures at acceptable levels in the Singapore weather over prolonged periods.

Unlike our original WAHRSIS version \cite{WAHRSIS} and commercial sky imagers, neither of the two new models uses a mechanical sun blocker. Instead, they rely on the fusion of multiple exposures to create a high-dynamic-range image and minimize over-saturation of pixels near the circumsolar region. Table \ref{tab:diff-WAHRSIS} summarizes the main improvements introduced by the two new models over the first.

\begin{table}[htb]
\small
\centering
\begin{tabular}{ |L|M|M|M| }
  \hline 
  WAHRSIS Properties  & \textbf{Previous model} &  \textbf{Sealed model} & \textbf{Ventilated model} \\
  
  \hline \hline
  
  Glare reduction method & Sun blocker & HDRI & HDRI  \\ \hline
  Occlusion severity & Moderate & Nil & Nil \\ \hline 
  Weatherability  & Limited & Good & Good \\ \hline
  Portability  & Limited & Good  & Good \\ \hline
  Image resolution  & High & High  & High \\ \hline 
  Total cost (in US\$)& \$2525  & \$1895 & \$1769  \\ \hline
\end{tabular}
\caption{Comparison of the different WAHRSIS models.}
\label{tab:diff-WAHRSIS}
\end{table}

We are using the captured images for sky/cloud segmentation and classification into different cloud types \cite{ICIP2015b}, and we are deploying multiple sky cameras in order to estimate cloud base height \cite{IGARSS2015b}. Our future work includes extending these methods to HDR images.

\section{Acknowledgments}
Darko Jokic (Hochschule f{\"u}r Technik Rapperswil, Switzerland) designed and built the sealed WAHRSIS model during his internship at NTU Singapore.  Charles Wong (NTU Singapore) designed and built the ventilated WAHRSIS model as part of his final year project.

\balance
\small 
\bibliographystyle{IEEEbib}

\end{document}